# DYNAMIC NETWORK UPDATING TECHNIQUES FOR DIAGNOSTIC REASONING

G.M.A. Provan
Computer and Information Science Department
University of Pennsylvania
Philadelphia  PA 19104-6389
phone:(215) 898-9830;  email: provan@cis.upenn.edu

## Abstract

A new probabilistic network construction system, DYNASTY, is proposed for diagnostic reasoning given variables whose probabilities change over time. Diagnostic reasoning is formulated as a sequential stochastic process, and is modeled using influence diagrams. Given a set $O$ of observations, DYNASTY creates an influence diagram in order to devise the best action given $O$. Sensitivity analyses are conducted to determine if the best network has been created, given the uncertainty in network parameters and topology. DYNASTY uses an equivalence class approach to provide decision thresholds for the sensitivity analysis. This equivalence-class approach to diagnostic reasoning differentiates diagnoses only if the required actions are different. A set of network-topology updating algorithms are proposed for dynamically updating the network when necessary.

## 1 INTRODUCTION

The development of graphical representations for probabilistic models (e.g. belief networks [Pearl, 1988], influence diagrams [Howard and Matheson, 1981; Shachter, 1986; Shachter, 1988]) has enabled efficient probabilistic models to be developed for many tasks, such as diagnostic reasoning [Pearl, 1988; Heckerman and Horvitz, 1990], natural language analysis[Goldman and Charniak, 1990], etc. These representations, by specifying the causal relationships among variables in a causal graph (and not all possible relationships), facilitate efficient inference. A great deal of the recent research in automated probabilistic reasoning has focused on developing more efficient and more general algorithms for causal probabilistic models, and on methods for incrementally constructing belief networks.

However, the application of these techniques and representations to complex diagnostic tasks, such as medical diagnosis, have oversimplified such tasks. A common simplification made in many current approaches is modeling the diagnostic process as a single-stage, static process. This is inadequate, as diagnostic reasoning is a sequential, dynamic process in which feedback is important. Provan and Poole [1991] point out the necessity of considering this complete process, and in particular, the effects of feedback.

This paper extends existing diagnostic models to incorporate the dynamic and sequential nature of diagnostic reasoning. It proposes techniques for constructing sequential belief networks, and of dynamically updating such networks. Many existing techniques for constructing belief networks (e.g. [Goldman and Charniak, 1990; Heckerman and Horvitz, 1990]) model the process for one instant of time.[1] For certain tasks this is adequate, but for tasks in which the probabilistic relationships among variables changes over time, it can be difficult to know when the best model has been constructed. This sometimes produces incorrect answers due to the selection of incorrect probabilities and/or causal relationships. Hence, both the diagnosis and the decision taken given this diagnosis may hinge on whether the best model has been constructed, given the data at a particular time $t$. Sensitivity analyses may be used to test how the data at different times affects the best decision. If the sensitivity analyses show that a better decision would be made under an alternative model, then the model needs to be updated. It is these sensitivity analyses and model updating techniques that are of interest here. Criteria are proposed to determine when network topology revisions are necessary given time-varying probabilistic and causal relationships. These criteria are based on examining the equivalence of outcomes (e.g. treatments for diseases). Algorithms for conducting the necessary revisions are outlined, including refinement and coarsening techniques [Chang and Fung, 1990], and other network

---

[1] This is true even for systems in which the models can be constructed incrementally, e.g. [Goldman and Charniak, 1990].

revision algorithms [Pearl, 1988; Srinivas and Breese, 1990].

This approach makes dynamic network updating possible, and formalizes the sequential nature of diagnostic reasoning (e.g. to allow feedback into the network). The explicit introduction of utilities into diagnostic models[2] allows a more realistic formalization of the diagnostic process. In addition, it is expected that the techniques developed for diagnostic reasoning may be applied to other domains, where appropriate.

## 2 DYNAMICS OF DIAGNOSTIC REASONING UNDER UNCERTAINTY

Treating a diagnostic task as being time-independent can lead to incorrect results in certain domains. Consider medical diagnosis, and in particular the diagnosis of abdominal pain. Constructing a model for the observation of abdominal pain should not be done for a single time interval, since, as noted in [Schwartz et al., 1986], many symptoms take on different meanings as diseases evolve over time, both in terms of their inter-relationships and the diseases indicated by the particular symptoms. In a possible case of appendicitis, the initial symptoms include non-specific abdominal pain (which could be confused with many other ailments), and are often accompanied soon thereafter by gastrointestinal distress and possibly by anorexia and fever. This pain subsequently becomes localized to the right lower quadrant (RLQ) of the abdomen (which then provides a strong indication of appendicitis, along with a high white blood count). If the appendix ruptures, then there are several more symptoms; however, a perforated appendix leads to serious internal complications.[3] Given the evolution of a disease such as appendicitis, the probabilities assigned to network nodes, and even the topology of the network itself, must change over time. For example, Figure 1 shows how the likelihood ratio for the diagnosis of appendicitis might change over time. Clearly, in the initial stages of appendicitis, many other diagnoses are equally likely given the symptoms.

A second aspect of this dynamic nature of (diagnostic) reasoning is the need for modeling the temporal order of observations. In some cases the temporal sequence of observations (as opposed to just an unordered list of the set of observations) can provide strong cues for a diagnosis. For example, if a woman has abdominal pain, noting whether this pain is immediately followed by gastrointestinal distress could help identify a pos-

---

[2] Utility considerations have been ignored in most formal models of diagnostic reasoning, except for approaches such as [Heckerman and Horvitz, 1990].

[3] Most diagnostic procedures attempt to avoid perforation and its resulting complications.

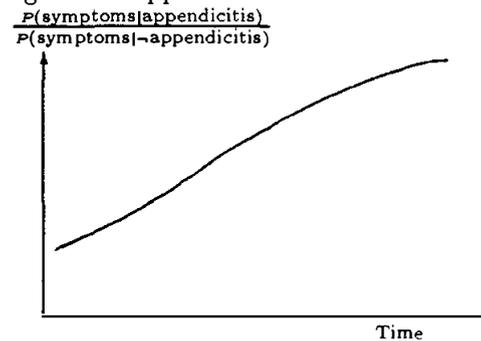

Figure 1: Change over time of likelihood ratio for the occurrence of Right-Lower-Quadrant pain given a diagnosis of appendicitis

$$\frac{P(\text{symptoms}|\text{appendicitis})}{P(\text{symptoms}|\neg\text{appendicitis})}$$

sible case of appendicitis, whereas the absence of such immediate distress would make the presence of a gonohorreal cyst in the right fallopian tube more likely. A second example is the diagnosis of a car which has trouble starting. The sequence of events leading to the inability to start can help identify the problem. Thus, the inability to start only on mornings after it has rained may indicate that moisture is getting under the distributor cap.

A third aspect is the ability to incorporate the effects of feedback. Feedback can alter not only the probability assignments to a network, but also the topology of the network. For example, consider a network constructed for a case of RLQ abdominal distress. If simple stomach upset is diagnosed, and a treatment of Diovol is administered, the persistence of RLQ abdominal distress will provide feedback to the system that the diagnosis may be incorrect, and the network topology and/or probabilities may need to be updated.

This paper proposes extensions to existing network construction techniques to model diagnostic reasoning as a sequential, dynamic process using the formalism of influence diagrams. This proposal is not intended to be a full temporal calculus based on Bayesian networks, as discussed in [Kanazawa, 1991], for example. Instead, it attempts to build simple networks which will realistically model the dynamics of diagnostic reasoning without necessitating the complicated (and computationally costly) construction and solution of temporal Bayesian networks.

## 3 SYSTEM ARCHITECTURE

There are many existing systems and theories for model construction. Examples of such network construction frameworks include the proposal of Lehmann [1990], and examples of such systems include QMR-DT [Shwe and Cooper, 1990] and FRAIL3 [Goldman and Charniak, 1990]. In each of these proposals, the goal is to construct a model which completely characterizes the data. However, this goal conflicts with the need for efficient performance of implemented systems. Solving Bayesian network models is NP-hard [Cooper, 1990], so the networks constructed must be as small as possible to ensure efficiency. The proposal presented in this paper trades off (to some extent) completeness and accuracy for efficiency, as is done in many other systems, such as [Heckerman and Horvitz, 1990].[4]

A new system architecture proposed to model dynamic reasoning tasks is depicted in Figure 2. This system is called DYNASTY, for DYnamic Network Analysis of System TopologY.

Like several existing network construction methods (e.g. QMR-DT, FRAIL3), we start with a Knowledge Base (KB) containing (1) causal rules, and (2) a set of conditional probability tables. From this KB a network is constructed to solve a given task.

The KB for DYNASTY consists of a network of nodes and arcs. Nodes represent state variables, and arcs exist between pairs of nodes related causally and/or temporally.

Associated with the network are probability tables for the conditional probabilities for the network, such as those required for the construction of a Bayesian network. In addition, utility values are stored for decision-making.

Typically, the complete KB for a given domain is quite large,[5] and given a set $O$ of observations, it is necessary to construct a network containing only the data related to $O$ (and not the entire KB).

Within the general model-construction framework (such as that described in Lehmann [1990]), there is always uncertainty in choosing the correct model. That uncertainty may be due to uncertainty in the instruments used to record data, to noise, or to the relationship between data from observations and causes for the observations (e.g. the diseases causing the observed symptoms). This paper examines the uncertainty arising from relating observations and causes, and in particular the temporal uncertainty of this relationship.

The remainder of the paper discusses the algorithms used to create an influence diagram from the KB, and for dynamically altering this influence diagram.

---

[4]The appropriate balance of resources between meta-analysis of model construction and model solution has been studied by [Horvitz et al., 1989; Breese and Horvitz, 1990].

[5]As an example, the QMR-DT network represents 534 diseases, 4040 manifestations and 40,740 disease-manifestation arcs [Heckerman and Horvitz, 1990].

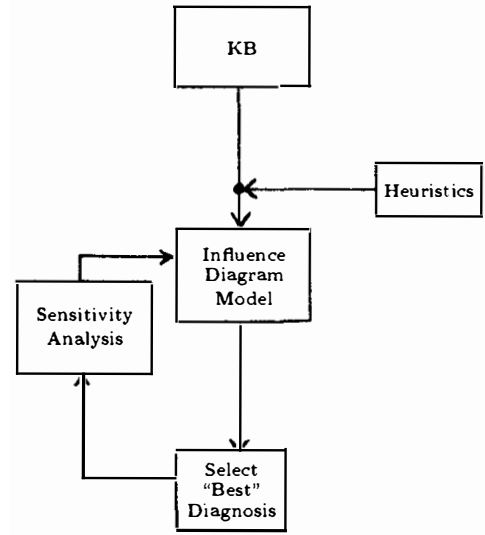

Figure 2: Network construction methods in DYNASTY

## 4  MODEL CONSTRUCTION HEURISTICS

### 4.1  Time Dependence

As noted earlier, diagnostic tasks whose characteristics change over time have not been modeled in earlier approaches. The approach taken in DYNASTY is to discretize the possible times from which the observations could have occurred. Call $\mathcal{D}_{t_i}$ the network (consisting of causes and intermediate causes/observations) which would need to be constructed at time $t_i$. In full generality, the networks at different times are different, and they can each be quite large for complicated tasks. To fully model a diagnostic task, an influence diagram (ID) containing sub-networks for each time $t_i$ would need to be constructed, given a set $O$ of observations. This is shown in Figure 3.

Figure 3: Most general influence diagram for solving a stochastic diagnostic task

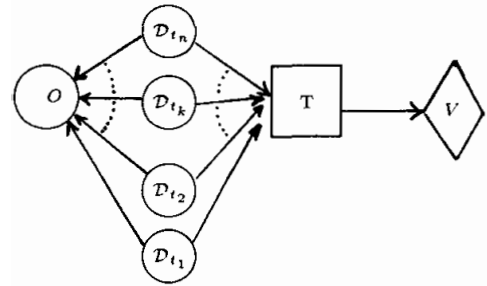

DYNASTY attempts to solve a simplified task: it creates a network for particular time $t_j$, and then conducts a sensitivity analysis to determine if the action taken is affected by the choice of time $t_j$. The ID which would be constructed is shown in Figure 4.

Figure 4: Simplified influence diagram for solving a stochastic diagnostic task

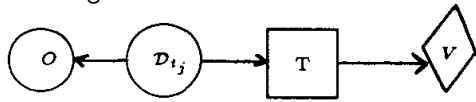

Example 1 Consider the time course of a possible case of appendicitis. Early in the course of appendicitis, the symptoms could appear to be a simple upset stomach. Figure 5 shows the notation necessary

Figure 5: Notation for constructing Abdominal Pain Influence Diagram

**OBSERVATIONS**   **HYPOTHESES**

```
α ≡ anorexia
N ≡ nausea              A  ≡ appendicitis
F ≡ fever               US ≡ upset stomach
P ≡ abdominal pain      FP ≡ food poisoning
LLQ ≡ LLQ pain          GC ≡ gonohorreal cyst
RLQ ≡ RLQ pain
```

to construct IDs for this task. If the observations are nausea and general abdominal pain, then the simple ID shown in Figure 6 may be constructed. This is an easy influence diagram to construct and solve. Given an ID such as this, the possible treatments are the administration of an emetic (for food poisoning) or Diovol (for simple upset stomach).

Figure 6: Simple influence diagram for abdominal pain example

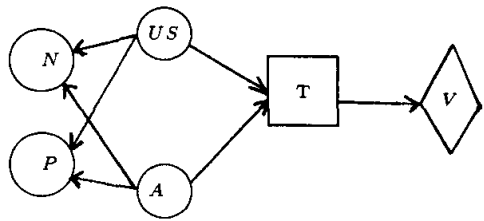

However, these observations may actually be indicative of the early stages of appendicitis. To make sure that a possible case of appendicitis might be diagnosed, the ID shown in Figure 7 must be constructed. This ID bears little relation to the ID shown in Figure 6. The possible treatments include: (1) emetic (for food poisoning), (2) Diovol (for simple upset stomach), (3) removal of appendix (for appendicitis), or (4) treatment or removal of gonohorreal cyst.

Figure 7: More complex influence diagram for abdominal pain example

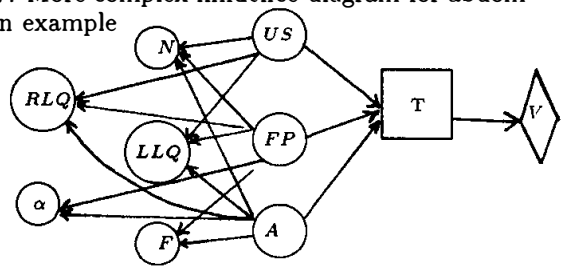

This example shows how, given a set of observations, uncertainty in the time course of possible diseases may require entirely different IDs. ♣

There are a number of heuristics used in DYNASTY for network construction. One heuristic is the use of temporal orderings for probability assignments. This heuristic is best demonstrated by an example. Consider the diagnosis of a car which infrequently has problems starting. The two diagnoses under consideration are a distributor cap problem (DC) or an alternator problem (ALT). The weather (W) may affect the diagnosis, as wet conditions can cause condensation under a distributor cap, thereby causing the failure of the car to start ($\overline{ST}$). Other possible causes of the problems in starting, e.g. the alternator may be faulty and not recharging the battery, are not affected by weather conditions. A simple Bayes network for this problem is shown in Figure 8. Knowledge of the history of the correlation between weather conditions and success in starting the car can significantly affect the probabilities assigned to the network. For example, if the car only gives trouble starting in wet conditions, then the problem is most likely DC; if the car gives trouble with equal probability in both wet and dry conditions, then the problem is most likely ALT. In fact, trouble in a single instance when the weather is dry will lead to the assignment of a low probability to $P(DC|\overline{ST}, W)$. In this case, the *history* of the problem is crucial to the probability assignment.

Figure 8: Bayesian network model for determining the cause of the failure of a car to start

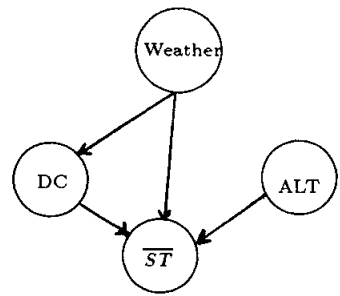

Hence, the *history heuristic* is the use of temporal history, whenever possible, in selecting the probabilities (from the probability tables) to be assigned to the network in consideration. The temporal history is computed simply by tracing the history for a node in the KB, using revised Truth Maintenance algorithms for computing the justifications for a node in a dependency network [McAllester, 1990]. The history heuristic also uses triggers to guide probability assignments. For example, finding a single instance when the car won't start in dry conditions is a trigger to the assignment of a low probability to $P(DC|\overline{ST},W)$.

### 4.2 Sequential Diagnostic Process

The ID framework also allows diagnostic reasoning to be formulated as a sequential diagnostic process. Using a result of Tatman and Shachter [1990], an ID can model a sequential process using dynamic programming, provided that the value function $V$ is separable. In terms of IDs, a value node is separable if it can be represented as the sum or product of multiple sub-value nodes.

Value node separability has been exploited in the design of a sequential process for image understanding [Levitt et al., 1990]. In a similar manner, value node separability is used to model the sequential nature of diagnostic reasoning. In brief, the decision nodes in a DYNASTY ID are called treatments, which may be tests to determine more observations, or actual treatments for hypothesized diseases. In the former case, given an ID shown in Figure 6, the test T can determine a new observation $O'$, creating a new ID with another decision node $T'$ (e.g. another test or a treatment) and another value node $V'$. In this manner, the sequential nature of tests (or treatments) providing feedback to the diagnostic process can be modeled.[6]

## 5  MODEL UPDATING

### 5.1  Overview

In a problem for which probabilities are temporally dependent, the sensitivity of the computed decisions to the temporally-dependent probabilities must be tested. This provides a threshold for determining when a better model is warranted. This may require new probability values (corresponding to a new time $t'$), or a new network topology corresponding to time $t'$.

This sensitivity analysis/model updating in DYNASTY occurs in two stages:

**Sensitivity Analysis** First, a sensitivity analysis is conducted to determine if data from time $t'$ provides a better model than the data from time $t$.

**Model Updating** If the network model needs to be updated, then some of the following processes may need to be invoked:

1. New probability values are assigned and propagated to compute a new network equilibrium state.
2. Network topology is altered.
3. A new model is built for a different time $t'$.

These processes are now discussed in greater detail.

### 5.2  Equivalence Class Sensitivity Analysis

Given the construction of an ID model at time $t$, a decision (with accompanying diagnosis) of maximal utility is computed. For example, in the car diagnosis example, the diagnosis might be DC, and the decision *REPLACE-DC*. This decision would maximise the requirement of ensuring that the car no longer has trouble starting.

In the process of computing this best decision, the next-best decision for a different equivalence class is also recorded. In the car example, this is *REPLACE-ALT*. If there is uncertainty concerning which probabilities are correct, then the sensitivity of the decision to this uncertainty must be determined. This is formalised in terms of equivalence classes of decisions as follows.

#### 5.2.1  Analysis of Equivalence Classes

The equivalence class approach to diagnosis, as originally formulated in [Provan and Poole, 1991], is summarised here. The rationale is that there is no point in distinguishing between decision-equivalent diagnoses, i.e. diagnoses for which the decision taken (e.g. administration of drugs to a patient) are the same; as far as the decision-maker is concerned decision-equivalent diagnoses should be considered as the same diagnosis.

The aim of diagnostic reasoning is to provide a treatment for a set of observations. From an equivalence-class point of view, this reduces to refining the set of use-equivalent possibilities; i.e. one does not care about distinct diagnoses, but *distinct treatments* (and their associated distinct equivalence classes). Thus, use-equivalence induces a partition on the set of diagnoses, where each partition corresponds to a possible distinct decision.

---
[6]Please refer to [Provan, 1991 (forthcoming)] for more details. The presentation here is brief due to space limitations.

Let $T$ be the set of all treatments (or decisions).[7] Let $\mathcal{D}$ be the set of all possible diagnoses.

**Definition 5.1** The **possible treatment space** $\mathcal{P}$ is a subset of $\mathcal{D} \times T$. $\langle D, T \rangle \in \mathcal{P}$ means that $T$ is a possible treatment given that the diagnosis is $D \in \mathcal{D}$.

$\mathcal{P}$ induces an equivalence relation on the set of diagnoses. This will be called *strong equivalence* with respect to $\mathcal{P}$. The idea is that equivalent diagnoses have the same set of possible treatments.[8]

**Definition 5.2** Two diagnoses $D_1$ and $D_2$ are **strongly equivalent** with respect to $\mathcal{P}$, written $D_1 \equiv_\mathcal{P} D_2$ if $\forall\, T \in \mathcal{T}$, $\langle D_1, T \rangle \in \mathcal{P}$ if and only if $\langle D_2, T \rangle \in \mathcal{P}$.

### 5.2.2 Equivalence Class Decision-making

We assume we have a measure $\mu(D, T)$ of the utility of treatment $T$ given diagnosis $D$. We can define the possible treatment space as the set of diagnoses with the same utility.[9] In this case, "strong use-equivalence" means having the same utility for each treatment.

Let $\mathcal{D}$ be the set of use-diagnoses. For $D \in \mathcal{D}$, every logical model of $D$ has the same utility measure. The following proposition about the expected value, $\mathcal{E}(T)$, of treatment $T$ was proven in [Provan and Poole, 1991]:

$$\mathcal{E}(T) = \sum_{D \in \mathcal{D}} \mu(D, T) \times p(D). \qquad (1)$$

Under this approach to diagnostic reasoning, diagnoses are selected such that the expected utility of the treatment is maximised. That is, the goal is to compute $\gamma_i$ such that the expected value of the treatment given by equation 1 is maximised.

Consider an ID in which the variables are denoted by $X = \{x_1, ...., x_n\}$, such that any diagnosis $D$ consists of a subset of variables $X' \subseteq X$ which are not functioning normally (cf. [de Kleer et al., 1990; Pearl, 1988; Provan and Poole, 1991] for a further description of such diagnostic models). Then equation 1 can be rewritten in terms of these variables as

$$\mathcal{E}[T] = \sum_{D \in \mathcal{D}} \sum_{D \models x} \mu(x, T) \times p(x), \qquad (2)$$

where $\mu(x, T)$ is the value of $\mu(D, T)$ such that $x$ is true in $D$.

The notion behind the sensitivity analysis is as follows: consider a model constructed at time $t$, such that decision $T_i$ is the optimal treatment. Call $\beta$ the expected utility for decision $T_i$. If the probabilities of certain variables are time-dependent, then these new probabilities need to be substituted into the model to check if the decision would change. Note that different diagnoses may be computed, but if the decision is unchanged, then, under this use-equivalent approach, no network updating is necessary. For network updating to be necessary, the threshold $\beta$ must be exceeded by the expected utility of another treatment $T_j$ given probabilities for time $t'$, i.e.

$$\left[ \mathcal{E}[T_j] = \sum_{D \in \mathcal{D}} \sum_{D \models x} \mu(x, T) \times p(x) \right] > \beta.$$

This provides a precise bound on when the treatment changes. When the threshold is exceeded, then network alterations may be necessary. These updating methods are now summarised.

## 5.3 Model Updating Techniques

There are several types of model updating operations, of which two of the most important are: (1) probability value updating, and (2) network topology updating. These are discussed in turn.

### 5.3.1 Probability Value Updating

This is the simple case of network updating. If no changes to the network topology are required when the model is updated from time $t$ to $t'$, then the required alterations to the probability values are made, and these values are propagated to obtain a new network equilibrium state.

For example, during the early stages of appendicitis diagnosis, probability values may need to be updated given changes in location of abdominal pain. Possible changes in probability assignments are shown in Figure 9(b),(c).

### 5.3.2 Network Topology Updating

Consider the onset of an entirely new set of symptoms in the observation of a patient with a possible case of the later stages of appendicitis. These are shown in Figure 7. If we started with the model in Figure 6, we see that the topology of the network needs to be altered.

---

[7] By a treatment we mean a total prescription of what to do (i.e., we do not conjoin different treatments — the conjunction would be one treatment). A treatment may be a test to distinguish abnormalities, the administration of drugs, replacement of circuit components, etc.

[8] Other types of equivalences, e.g. weak equivalence, are also distinguished in [Provan and Poole, 1991]; such cases are not discussed here due to space limitations.

[9] Formally, the treatment in the possible treatment space would be a pair $\langle T, v \rangle$ where $\langle D, \langle T, v \rangle \rangle \in \mathcal{P}$ if $\mu(D, T) = v$.

Figure 9: Early stages of the diagnosis of appendicitis

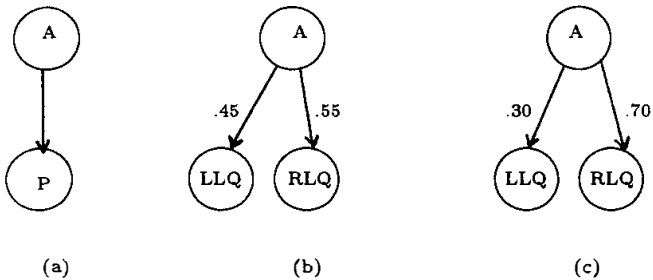

(a)           (b)           (c)

If changes to the network topology are required when the model is updated from time $t$ to $t'$, then one of several algorithms may be used. These algorithms include:

**Refinement/coarsening** Refinement/coarsening operations [Chang and Fung, 1990] are used to split/merge network nodes respectively. Consider a network refinement necessary to include new alternatives. For example, in abdominal diagnosis, the construction of a network which models only lower abdominal pain may need to be refined to differentiate right-lower quadrant (RLQ) and left-lower quadrant (LLQ) pain. Hence, a node modeling lower abdominal pain needs to be split into nodes for RLQ and LLQ (cf. Figures 9(a),(b)). Or in the car diagnosis example, the single node for weather may need to be split into nodes for wet weather and mixed (wet and dry) weather.

The network changes made for the refinement/coarsening operations are local, and do not involve all nodes in the network. This is formalised as follows. If $x$ is a state node, then we call $\Pi_x$ the predecessors of $x$ in the network, and $\Sigma_x$ the successors of $x$ in the network. The Markov boundary of $x$ is the minimal set of nodes which "shield" $x$ from the rest of the network. The Markov boundary $\mathcal{M}(x)$ of node $x$ consists of $\Pi_x \cup \Sigma_x \cup \Pi_{\Sigma_x}$. Hence, ensuring the joint probability distribution of $\mathcal{M}(x)$ is unaffected by the refinement/coarsening or $x$ ensures that the rest of the network will be unaffected as well.

For example, it is shown in [Chang and Fung, 1990] that in a refinement of the values of the state space of variable $x$, $\Omega_x$, each value $\omega_x \in \Omega_x$ is refined into multiple values $\omega'_x \in R(\omega_x)$. For each value $\omega_x \in \Omega_x$ which is refined into a value $\omega'_x \in R(\omega_x)$,

$$p(\Sigma_x|\omega_x, \Pi_{\Sigma_x})p(\omega_x, \Pi_x)$$
$$= \sum_{\omega'_x \in R(\omega_x)} p(\Sigma_x|\omega'_x, \Pi_{\Sigma_x})p(\omega'_x, \Pi_x) \quad (3)$$

must be satisfied for all values of $\Pi_x$. This provides a set of constraints on how $\mathcal{M}(x)$ must be altered. In an analogous manner, constraints can be defined for the coarsening of the values of the state space of variable $x$, $\Omega_x$, where multiple values of $\omega_x \in \Omega_x$ are combined into a single value $\omega'_x \in C(\omega_x)$.

The coarsening operation is defined similarly [Chang and Fung, 1990]. The coarsening operation may lose information during the process of node aggregation (i.e. the network probability assignments may be altered). Using the equivalence-class approach, such information loss is acceptable if the equivalence class does not change. Otherwise, approximations may need to be used [Chang and Fung, 1990].

**Network additions** Instead of splitting and/or merging existing nodes, completely new nodes may need to be added to, or particular nodes deleted from, the network. In such cases a variety of other algorithms are invoked, such as the reduction and clustering algorithms present in the IDEAL system algorithm library [Srinivas and Breese, 1990]. In network addition, the KB is consulted to determine which nodes must be added based on causal relationships.

**Network Re-instantiation** It may turn out that the network created is inappropriate for the diagnostic task. For example, a simple network may be created which cannot be appropriately augmented to model a more complicated case.[10] In such a situation, a completely new network is constructed from the KB.

### 5.4 Implementation

The KB is implemented in Common Lisp. Extended Justification-based TMS (e.g. [McAllester, 1990]) data structures and algorithms are used for determining relevant nodes to instantiate given a set of observations. The influence diagrams are implemented using the IDEAL system [Srinivas and Breese, 1990].

It is hoped that the TraumAID system [Webber et al., 1990] will be used as a test-bed for this system. TraumAID is a decision support tool for the management of multiple trauma. Trauma management includes both diagnosis and treatment, and this diagnostic tool achieves these features using two modules: (1) a rule-based reasoner which models the relationships between clinical evidence and diagnostic/therapeutic goals, and (2) a planner which manages the achievement of multiple goals. TraumAID is an excellent system on which to test the theoretical results because, unlike most similar systems, it already contains a no-tion of sequential action and change, key elements of the proposed theory of diagnostic reasoning. Further, efficient incremental management of action and change is necessary for trauma management.

---

[10]If radical changes must be made to an initial network, it can be computationally cheaper to create a new network from scratch than to alter the original network using coarsening/refinement operations.

## 6 CONCLUSIONS

This paper has described a proposed dynamic network construction system which can build models for problems with temporally-dependent probabilities. Heuristics are used to identify the best possible model, and to test the sensitivity of this model to probability values over time. Given the network updating capabilities of DYNASTY, the full diagnostic cycle, which includes feedback from the decisions made, can be incorporated into the network. In addition, the ability to refine/coarsen the network enables different levels of granularity (i.e. the coarseness of the description of the system being modeled) to be examined during the diagnostic process. Most other approaches to diagnostic reasoning (e.g. [de Kleer et al., 1990]) have no way of dynamically altering the granularity of the system description.

Future work includes testing the feasibility of the algorithms in DYNASTY on real-world problems, and extending and optimising these algorithms. The KB for the TraumAID system is the first set of real data for which such tests are proposed.

**ACKNOWLEDGEMENTS:** The comments of the anonymous reviewers have led to improvements in the paper.